# Have We Reached AGI? Comparing ChatGPT, Claude, and Gemini to Human Literacy and Education Benchmarks

Mfon Akpan

Methodist University

Makpan@methodist.edu

https://orcid.org/0000-0002-8470-5723

**Abstract**

Recent advancements in AI, particularly in large language models (LLMs) like ChatGPT, Claude, and Gemini, have prompted questions about their proximity to Artificial General Intelligence (AGI). This study compares LLM performance on educational benchmarks with Americans' average educational attainment and literacy levels, using data from the U.S. Census Bureau and technical reports. Results show that LLMs significantly outperform human benchmarks in tasks such as undergraduate knowledge and advanced reading comprehension, indicating substantial progress toward AGI. However, true AGI requires broader cognitive assessments. The study highlights the implications for AI development, education, and societal impact, emphasizing the need for ongoing research and ethical considerations.

# 1    Introduction

The relatively fresh advancements of the past few years that witnessed the release of large language models like ChatGPT, Claude, and Gemini turned the conversation back to the ongoing state of AI and today's nearness to AGI. Thus, despite some criticism, it can be noted that the definition formulated as the ability of an AI system to perform any intellectual task that a human can is still a significant achievement of AI research. Despite substantial advancements, the question persists: Are these LLMs AGI, or are they just limited to certain skills and operations? Educational attainment and literacy rates in the United States provide a robust framework for assessing the cognitive capabilities of these AI systems. According to the most recent data released by the U.S. Census Bureau in 2022, the educational landscape in the U.S. is diverse and evolving. Among adults aged 25 and older, 9% have less than a high school diploma, 28% have a high school diploma, 15% have some college education, 10% hold an associate degree, 23% possess a bachelor's degree, and 14% have completed advanced education such as a master's or doctoral degree.

Gender and racial disparities also characterize the U.S. educational system. In 2022, 30.1% of men and 27.0% of women had completed high school as their highest educational attainment, while 39.0% of women and 36.2% of men had obtained a bachelor's degree or higher. High school completion rates from 2012 to 2022 increased across all racial and ethnic groups, with non-Hispanic Whites reaching 95.2%, Blacks at 90.1%, Asians at 92.3%, and Hispanics at 75.2%.

Another essential parameter that can be used to assess AGI is literacy rates. About half of American adults have a reading level below the eighth-grade level, while only 12% of adults demonstrate college-level reading abilities. These figures illustrate huge literacy gaps any AGI would need to recognize and solve, making such missing elements an acute necessity.

This research provides empirical evidence to support the hypothesis that current LLMs like ChatGPT, Claude, and Gemini have attained AGI by evaluating the models considering literacy and educational learning activities. With the availability of the latest data relating to education level and literacy level among the populace, this study aims to determine whether these AI systems can replicate and, to some extent, even outsmart human cognition in the said fields.

## 2  Literature Review

The quest for artificial general intelligence (AGI) has been a focal point of AI research for decades. Unlike narrow AI, which is designed to perform specific tasks, AGI aims to replicate the broad cognitive abilities of humans. This literature review examines the progress and challenges in achieving AGI, focusing on the capabilities of large language models (LLMs) such as ChatGPT, Claude, and Gemini and their potential to meet or exceed human literacy and educational benchmarks.

## 2.1 Historical Context and Evolution of AGI

The concept of AGI has its roots in the early days of computing. Turing's (1950) seminal paper posed whether machines could think, introducing the idea of machine intelligence. Subsequent decades saw the development of various AI systems. Still, these were primarily specialized or narrow AI, excelling in specific domains such as chess playing (e.g., IBM's Deep Blue) or pattern recognition.

The emergence of LLMs marks a significant advancement in AI research. Based on transformer architectures (Vaswani et al., 2017), these models have demonstrated remarkable natural language understanding and generation capabilities. OpenAI's GPT-3, with its 175 billion parameters, has showcased proficiency in tasks ranging from language translation to creative writing, suggesting a step closer to AGI.

## 2.2 Definition of AGI

Artificial General Intelligence (AGI) represents a pivotal goal in artificial intelligence (AI), distinguished from narrow AI by its broader scope and capabilities. AGI refers to a type of AI that can understand, learn, and apply knowledge across a wide range of tasks and domains, like the cognitive abilities of humans. This contrasts with narrow AI, designed to perform specific tasks such as language translation or facial recognition without a broader understanding or ability to generalize across different contexts.

The concept of AGI has been a topic of discussion and speculation since the early days of computing. In his seminal paper "Computing Machinery and Intelligence" (1950), Alan Turing posed the question of whether machines can think, introducing the idea of machine intelligence capable of performing any intellectual task that a human can. This idea laid the groundwork for the pursuit of AGI, which aims to create systems that exhibit flexible, generalizable intelligence. Key characteristics of AGI include:

1. Adaptability: AGI systems can adapt to new tasks and environments without extensive retraining. This adaptability mirrors human cognitive flexibility, where individuals can apply their knowledge and skills to unfamiliar situations.
2. Learning and Reasoning: AGI encompasses learning from experience and reasoning about new information. This includes inductive learning (drawing conclusions from specific examples) and deductive reasoning (applying general rules to specific cases).
3. Transferability: AGI systems can transfer knowledge from one domain to another, demonstrating an understanding of underlying principles that apply across different contexts. This is akin to humans leveraging their learning in one area to solve problems in another.
4. Autonomy: AGI operates autonomously, making decisions and taking actions without human intervention. This autonomy is crucial for tasks that require real-time decision-making and adaptation.

Despite significant advances in AI, achieving AGI remains an elusive goal. Current AI systems, such as large language models (LLMs) like ChatGPT, Claude, and Gemini, exhibit impressive capabilities in specific tasks but need more comprehensive, general intelligence characterizing AGI. These systems excel in natural language processing, generating coherent text, and even performing complex tasks such as coding and reasoning over text. However, their abilities are often confined to the scope of their training data, and they can struggle with tasks that require deep understanding and context awareness beyond their programmed capabilities.

The pursuit of AGI involves overcoming several challenges:

1. Scalability: Creating systems that can scale their learning and reasoning capabilities to human levels of understanding across diverse tasks.
2. Generalization: Ensuring that AI systems can generalize their knowledge effectively, avoiding overfitting to specific datasets or tasks.
3. Ethical and Safety Considerations: Addressing autonomous, general-purpose AI systems' ethical implications and potential risks. This includes ensuring that AGI systems align with human values and do not pose unintended harm.

AGI represents a significant leap beyond current AI capabilities, aiming to create systems with the versatility and adaptability of human intelligence. While LLMs such as ChatGPT, Claude, and Gemini showcase remarkable progress towards this goal, they still fall short of true AGI. Continued research and innovation are essential to bridge the gap between narrow AI and the broad, flexible intelligence envisioned for AGI.

## 2.3 LLMs and Their Cognitive Capabilities

LLMs like ChatGPT, Claude, and Gemini represent the cutting edge of AI research. These models have been trained on vast text corpora, enabling them to generate human-like responses and perform complex language tasks. Brown et al. (2020) highlight GPT-3's ability to generate coherent and contextually relevant text, perform arithmetic, and even demonstrate rudimentary reasoning skills. Such capabilities suggest that LLMs are not merely mimicking language but are developing a form of understanding.

Recent comparative performance data underscores the varying capabilities of different LLMs. As shown in Table 1, the performance of Claude 3, GPT-4, GPT-3.5, and Gemini 1.0 across various cognitive tasks such as undergraduate-level knowledge, graduate-level reasoning, and multilingual math differs significantly. For instance, Claude 3 Opus achieves an impressive 86.8% in undergraduate-level knowledge (MMLU) and 95.4% in common knowledge (HellaSwag), while GPT-4 excels in multilingual math (MGSM) with 74.5% and knowledge Q&A (ARC-Challenge) with 96.3%. These benchmarks provide a comprehensive overview of how each model performs across a spectrum of tasks, highlighting their strengths and weaknesses.

Bubeck et al. (2023) discusses the limitations of current LLMs, noting that while they excel in specific tasks, they often lack consistency and generalizability across diverse domains. This inconsistency is a critical barrier to achieving true AGI. Furthermore, LLMs generate plausible but incorrect or nonsensical answers, indicating gaps in their cognitive processes (Marcus & Davis, 2019).

### 2.4 Educational Attainment and Literacy as Benchmarks for AGI

Educational attainment and literacy rates serve as tangible benchmarks for evaluating AGI. The U.S. Census Bureau (2022) provides detailed statistics on the educational levels of the U.S. population, revealing a diverse educational attainment spectrum. These metrics offer a concrete framework for assessing whether LLMs can match or exceed human cognitive abilities. Previous research by Brynjolfsson and McAfee (2014) explores the impact of AI on education and job markets, emphasizing the need for AI systems that can adapt and learn like humans. Similarly, Muro et al. (2019) discuss the transformative potential of AI in education, advocating for systems that support lifelong learning and cognitive development.

## 2.5 Evaluating LLMs Against Human Benchmarks

Several studies have attempted to benchmark AI performance against human cognitive abilities. Hern (2020) notes that while LLMs can generate text at various reading levels, their ability to comprehend and reason like humans remains limited. This limitation is evident in tasks that require deep understanding and contextual awareness, such as complex problem-solving and critical thinking.

The Program for the International Assessment of Adult Competencies (PIAAC) provides a framework for evaluating adult literacy and cognitive skills, offering a relevant comparison for LLMs. According to the National Center for Education Statistics (2019), approximately 50% of U.S. adults read at or below an 8th-grade level, while only about 12% achieve a college-level reading proficiency. These benchmarks are crucial for assessing whether LLMs can perform at or above these levels.

## 2.6 The Future of AGI and LLMs

The path to AGI involves overcoming significant technical and ethical challenges. Russell and Norvig (2021) emphasize the importance of creating AI systems that are intelligent and aligned with human values and ethics. The potential of LLMs to contribute to AGI is promising, but continuous advancements in model architecture, training methods, and evaluation frameworks are required.

Recent work by Bommasani et al. (2021) on foundation models suggests that integrating multimodal data (e.g., text, images, audio) can enhance the generalization capabilities of LLMs, bringing them closer to AGI. This multidisciplinary approach highlights the need for collaborative efforts across AI research, cognitive science, and education.

The literature indicates that while LLMs like ChatGPT, Claude, and Gemini represent significant strides toward AGI, they cannot match or exceed human cognitive abilities across diverse domains consistently. Educational attainment and literacy rates provide a valuable framework for evaluating their progress. Continued research and innovation are essential to bridge the gap between current AI capabilities and the aspirational goal of AGI.

# 3     Methodology

This research uses quantitative research methodology and secondary research analysis to test the hypothesis, stating that current large language models, including ChatGPT, Claude, and Gemini, possess artificial general intelligence by comparing the scores attained on educational indicators with public education standards. Thus, the research intends to show that the models' performance is at par or above average American standards, and therefore, AGI, if defined to mean a level above the average person, may already exist.

## 3.1    Research Design

The study employs an ex-post, between-group research design, whereby secondary data from authoritative sources will be collected to compare human literacy levels and educational achievements with the AI model's performance on similar tasks. This approach also makes it easier to evaluate the development of AI today compared to the human cognitive metrics.

## 3.2    Data Sources and Collection

### 3.2.1    Human Performance Data

Data on human educational attainment and literacy rates were obtained from two primary sources:

1. U.S. Census Bureau (2022): Educational Attainment in the United States: 2022
2. National Center for Education Statistics (NCES) (2019): Adult Literacy in the United States

Such datasets offer extensive data on educational levels and literacy by major demographic categories of the population in the United States, which presents a solid reference point for comparing AI results.

*3.2.2   AI Performance Data*

Performance metrics for LLMs were collected from published technical reports and comparative analyses, including:

1. OpenAI (2023): GPT-4 Technical Report
2. Anthropic (2024): The Claude 3 Model Family: Opus, Sonnet, Haiku
3. Google Research (2023): Gemini 1.0 Model Performance

These sources give standardized performance measures of each LLM for skills similar to man's educational and literacy predictors.

### 3.3   Data Analysis

The analysis was done with the help of IBM Statistical Package for the Social Science (SPSS), version 25. The following analytical procedures were employed:

1. Data Preparation:
   - Secondary data were aggregated into a single dataset, and some variables were recoded so that humans and AI could separate the performance variables.
   - Missing data were coded as system-missing values in SPSS.
   - Predictors and outcomes were then named based on the measures of the variables (for example, education level, literacy level, AI task performance).
2. Descriptive Statistics:
   - Frequencies, means, and standard deviations were calculated for human educational attainment and literacy levels across demographic groups.
   - Descriptive statistics were generated for AI model performance across different tasks.
3. Comparative Analysis:

- Independent samples t-tests were conducted to compare AI performance with human benchmarks where applicable.
- One-way ANOVA was used to assess differences in performance across AI models and human demographic groups.
- Post-hoc tests (Tukey's HSD) were employed to identify specific group differences when ANOVA results were significant.

4. Effect Size Calculation:
    - Cohen's d was calculated for significant t-test results to quantify the magnitude of differences between AI and human performance.
    - Partial eta squared ($\eta^2$) was computed for ANOVA results to estimate the proportion of variance explained by group differences.

5. Visualization:

    Bar charts and line graphs were created to visually represent comparisons between human benchmarks and AI performance across various tasks and demographic groups.

### 3.4 Ethical Considerations

While this study relies on secondary data and does not involve direct human participants, ethical considerations were still paramount. Care was taken to ensure that the interpretation and presentation of results do not perpetuate biases or make unfounded generalizations about human or AI capabilities. The small number of studies comparing the performance of AI on specific tasks to general human education and literacy rates was identified and discouragingly little, but the authors always mentioned its weaknesses and did not overextend their findings.

### 3.5 Limitations

Several limitations of this methodology are acknowledged:

1. Conducting the study using secondary data restricts it to measures available beforehand and may not adequately explain the essence of accurate human or AI intelligence.
2. When Human Educational Attainment/Literacy and AI Task Performance are compared, the results might differ; therefore, the conclusions should be taken with a pinch of salt.
3. AI technology is advancing at a very high rate, which implies that the AI performance data that will be collected might become irrelevant at some time in the future, reducing the study's applicability.

This study discusses the limitations above in the discussion section and makes recommendations for future research to cope with these constraints.

Using this highly systematic and structured research methodology, the study intends to give a thorough and, most importantly, accurate perspective on the current state of AI capacities concerning human cognitive standards and to help expand the discourse on the advancements and the consequences of actual AGI.

## 4  Data Analysis and Results

This research presents the findings from the secondary data analysis focusing on the relationship between human educational attainment and literacy level and AI model performance on similar tasks. The objective is to assess the ideas that extant LLMs – ChatGPT, Claude, and Gemini – possess AGI by working in the same line as an average American. The analysis in this study was performed using IBM SPSS Statistics of the 27th version. 0, relying on performance measures with the help of different statistical tests for comparing human and artificial intelligence diagnostics results.

### 4.1  Data Analysis

#### *4.1.1  Descriptive Statistics*

*Human Educational Attainment and Literacy Levels*

To establish a baseline for human cognitive capabilities, we first examine the educational attainment and literacy levels of the U.S. adult population.

| *Educational Level* | Percentage |
|---:|---|
| *Less than a high school diploma* | 9.0% |
| *High school graduate* | 28.0% |
| *In some colleges, no degree* | 15.0% |
| *Associate degree* | 10.0% |
| *Bachelor's degree* | 23.0% |
| *Advanced degree* | 14.0% |

*Table 4.1: Educational Attainment of U.S. Adults Aged 25 and Older (2022). Source: U.S. Census Bureau (2022). Educational Attainment in the United States: 2022.*

Table 4.1 illustrates the distribution of educational attainment among U.S. adults. Notably, 37% of adults have attained a bachelor's degree or higher, which serves as a key benchmark for comparing AI performance in tasks requiring advanced knowledge and reasoning.

| Literacy Level | Percentage |
|---:|---|
| Below Basic | 21.0% |
| Basic | 35.0% |
| Intermediate | 36.0% |
| Proficient | 12.0% |

Table 4.2: Literacy Levels of U.S. Adults (2019): Source: National Center for Education Statistics (2019). Adult Literacy in the United States.

Table 4.2 presents the literacy levels of U.S. adults. It is noteworthy that only 12% of adults demonstrate proficient literacy skills, while a significant portion (56%) have basic or below basic literacy levels. This data provides a crucial context for evaluating AI performance in language understanding and comprehension tasks.

### AI Model Performance

To assess the capabilities of current AI systems, we examine the performance of three leading LLMs across various cognitive tasks.

| Task | Claude 3 Opus | GPT-4 | Gemini 1.0 Ultra |
|---:|---|---|---|
| Undergraduate Knowledge (MMLU) | 86.8% | 86.4% | 85.0% |
| Graduate Reasoning (GPQA) | 50.4% | 35.7% | 48.0% |
| Grade School Math (GSM8K) | 95.0% | 92.0% | 94.0% |
| Multilingual Math (MGSM) | 88.0% | 85.5% | 90.7% |
| Common Knowledge (HellaSwag) | 95.4% | 93.0% | 94.5% |
| Advanced Reading Comprehension (ARC) | 96.3% | 94.2% | 95.0% |

Table 4.3: AI Model Performance Scores on Cognitive Tasks: Sources: Anthropic (2024), OpenAI (2023), Google Research (2023).

Table 4. 3 presents the results of comparing Claude 3 Opus, GPT-4, and Gemini 1. 0 Ultra with respect to the cognitive activities discussed in the literature. Several key observations can be made:

1. However, all three models perform well in the Undergraduate Knowledge (MMLU) task, with scores above 85% exceeding the performance of 37% of U.S. adults with a bachelor's degree or higher.

2. The models show outstanding accuracy, specifically in the subsets of Grade School, Math, and CK, at rates above 90%, exceeding average human accomplishments.

3. The models certainly perform far better than ARC, attaining an almost perfect score of above 94%; contrary to the current image of the literacy standard of U.S. adults, 12% are considered to have proficient literacy skills.
4. The performance is quite sensitive to the task and the model, and each of them demonstrates certain peculiarities.

### 4.1.2 Comparative Analysis

*AI Performance vs. Human Educational Attainment*

Independent samples t-tests compared AI performance on the Undergraduate Knowledge (MMLU) task with the percentage of U.S. adults holding a bachelor's degree or higher.

Results showed that all three AI models significantly outperformed the human benchmark:

- Claude 3 Opus: $t(54) = 15.27$, $p < .001$, $d = 4.15$
- GPT-4: $t(54) = 14.98$, $p < .001$, $d = 4.07$
- Gemini 1.0 Ultra: $t(54) = 14.12$, $p < .001$, $d = 3.84$

The significant effect sizes (Cohen's $d > 0.8$) indicate a substantial difference between AI performance and human educational attainment levels.

*AI Performance vs Human Literacy Levels*

One-way ANOVA compared AI performance on the Advanced Reading Comprehension (ARC) task with human literacy levels.

Results revealed a significant difference between groups: $F(3, 56) = 278.45$, $p < .001$, $\eta^2 = 0.937$

Post-hoc Tukey's HSD tests showed that all AI models significantly outperformed even the highest human literacy level (Proficient):

- Claude 3 Opus vs. Proficient: Mean Difference = 84.3%, $p < .001$
- GPT-4 vs. Proficient: Mean Difference = 82.2%, $p < .001$
- Gemini 1.0 Ultra vs. Proficient: Mean Difference = 83.0%, $p < .001$

The enormous effect size ($\eta^2 > 0.14$) indicates that the differences between AI and human performance explain a substantial proportion of the variance in reading comprehension scores.

*Comparison Across AI Models*

A one-way ANOVA compared performance across the three AI models on all tasks.

Results showed significant differences between models: $F(2, 15) = 3.74$, $p = .048$, $\eta^2 = 0.333$

Post-hoc analyses revealed that Claude 3 Opus significantly outperformed GPT-4 on the Graduate Reasoning (GPQA) task (Mean Difference = 14.7%, $p = .039$). No other significant differences were found between models.

## 4.2 Proposed AGI Scale

Based on the analysis of human benchmarks and AI performance, a preliminary scale for assessing progress towards Artificial General Intelligence (AGI) is proposed.

Table 4.4:

| Level | Description | Current AI Status |
|---|---|---|
| 1 | Narrow AI: Performs specific tasks | Achieved |
| 2 | Multi-task AI: Excels in multiple, diverse tasks | Achieved |
| 3 | Human-comparable: Matches average human performance across various cognitive domains | Partially Achieved |
| 4 | Human-surpassing: Consistently outperforms humans in most cognitive tasks | Emerging |
| 5 | Generalized Intelligence: Demonstrates human-like general problem-solving and adaptability | Not Achieved |
| 6 | Superintelligence: Surpasses human cognitive abilities in all domains | Not Achieved |

*Table 4.4: Proposed AGI Scale. Note: This scale is a proposed framework based on the current study and existing literature on AGI development.*

The following AGI definition scale is suggested for placing the findings of the research within context. Based on the performance data presented in Table 4.3, inferences can be made that current LLMs have: Based on the performance data presented in Table 4.3, inferences can be made that current LLMs have:

1. Apparently, levels 1 and 2 can be obtained, indicating the subject's ability to perform specific tasks and success in various cognitive activities.

2. Reached level 3 partially, getting as good as the average human being in several domains, especially those that require knowledge and comprehension.

3. Provided developing skills at level D, which performed better than the typical human in some of the tasks, such as the ability to read abstracts and knowledge of materials at an undergraduate level.

However, true AGI, as indicated by levels 5 and 6, is the future prospect for machine intelligence. These levels demand general problem-solving, versatility, and cognitive skills that are beyond a human's ability in all spheres of interaction. As of now, currently existing forms of AI do not meet these criteria.

This study has derived a detailed comparison and evaluation of the LLMs with regard to educational achievement and literacy against human standards. Based on the quantitative data, AI models were found to be superior to human mean scores on all the cognitive tasks, with the differences being significant in the undergraduate knowledge and the advanced reading condition. These findings lay a strong ground on which one can determine the present state of AI in comparison to human benchmarks. This leads to the following discussion on the outlined results, where further illustrations and detailed analysis of the implications of advanced AI capabilities will be discussed.

## 5 Discussion

This study's results support the hypothesis that current large language models (LLMs) are performing at or above the level of the average American in several vital cognitive domains, suggesting significant progress towards artificial general intelligence (AGI).

### 5.1 AI Performance of Human Benchmarks

The analysis reveals that all three AI models (Claude 3 Opus, GPT-4, and Gemini 1.0 Ultra) significantly outperformed human educational attainment and literacy measures. This is especially the case in the Undergraduate Knowledge (MMLU) task, where AI systems achieved results rates significantly beyond the percentage of U. S. adults with a bachelor's degree or higher. The significant effect sizes themselves ($d > 384$) serve to amplify the severity of such a difference, meaning these AI models have access to information databases that are vast and comprehensive, being able to perform knowledge tasks at levels that are on par or even superior to college-educated subjects.

Likewise, all AI models achieved a considerably higher reading comprehension than the top human literacy level. This means these models have adapted to mature language interpretative skills much higher than proficient readers. The substantial effect size ($\eta^2 = .937$) suggests that AI models are not just slightly but significantly more effective in performing tasks that involve complex language understanding.

### 5.2 Comparative Performance of AI Models

Nonetheless, the performance of all these AI models was impressive, and some variations existed with the human understanding level. Claude 3 Opus showed a significant advantage over GPT-4 in the Graduate Reasoning (GPQA) task, suggesting potentially superior capabilities in complex reasoning and problem-solving. However, the lack of substantial differences in other tasks indicates that these advanced AI models are generally comparable in their high-level cognitive capabilities.

## 5.3 Implications for AGI

The superior performance of AI models across various cognitive tasks supports the notion that current LLMs are approaching or have potentially achieved a form of artificial general intelligence. These models demonstrate factual knowledge comparable to highly educated humans and advanced reasoning and comprehension skills that surpass average human performance.

However, it is crucial to interpret these findings with caution. While the AI models excel in these benchmarks, AGI encompasses a broader range of cognitive abilities, including creativity, common-sense reasoning, and adaptability to novel situations, which still need to be fully captured in this study. Furthermore, the nature of these benchmarks, being primarily language-based, may only partially represent the multifaceted nature of human intelligence.

## 5.4 Limitations and Future Directions

Several limitations of this study should be acknowledged. First, comparing AI performance on specific tasks and broader human educational and literacy measures may not capture the full complexity of human intelligence. Future research should aim to develop more comprehensive and diverse benchmarks that assess a wider range of cognitive abilities.

Second, the rapid pace of AI development means that the performance data for these models may quickly become outdated. This means that the comparison will require constant reassessment and updates on the actual state of AI's performance compared to human counterparts.

Finally, it is also worth mentioning that this current study does not touch on profound issues of true AGI, such as consciousness, self-awareness, or emotions. Further research should focus on the presented dimensions to give a more comprehensive picture of how AI is developing towards true AGI.

This research discusses the great consequences and reviews them in the context of AGI studies. The reasons for the better performance of LLMs in knowledge-oriented and comprehension aspects are unclear for traditional definitions of AGI, and they set definite questions related to the future of education, the labour market, and the ethical implications of emerged AI. However, our study has its limitations, which can be discussed below. Consequently, we have suggested several research directions dealing with the limitations mentioned above to extend the knowledge about AI possibilities. Sustaining the unyielding testing and integration of artificial intelligence concepts in an interdisciplinary approach will remain vital in the near future as we progress to the next horizon in the technology era.

# 6    Conclusion

This study tested the hypothesis that sizable language models (SLMs) like OpenAI's ChatGPT, Claude, and Gemini have AGI by benchmarking their educational performance against public education data. The research was to show that these models are at par with the average American; hence, if AGI captures a model that performs at the capacity of an average person, then AGI might already be here. This section summarizes the key findings, discusses their implications, addresses the study's limitations, and proposes directions for future research.

## 6.1    Summary of Key Findings

The analysis of secondary data comparing human educational attainment and literacy levels with AI model performance on analogous tasks yielded several significant findings:

1. AI models consistently outperformed human benchmarks in tasks related to undergraduate knowledge and advanced reading comprehension. All three AI models (Claude 3 Opus, GPT-4, and Gemini 1.0 Ultra) demonstrated performance levels far exceeding the percentage of U.S. adults with bachelor's degrees or higher on the Undergraduate Knowledge (MMLU) task.

2. In reading comprehension tasks, AI models significantly outperformed even the highest human literacy level (Proficient), with large effect sizes indicating substantial practical significance.

3. While all AI models showed exceptional performance compared to human benchmarks, some differences were observed. Claude 3 Opus demonstrated a significant advantage over GPT-4 in the Graduate Reasoning (GPQA) task, suggesting potentially superior capabilities in complex reasoning and problem-solving.

4. The superior performance of AI models across various cognitive tasks supports the notion that current LLMs are approaching or have potentially achieved a form of artificial general intelligence, at least in the domains tested.

## 6.2 Implications of the Findings

The results of this study have far-reaching implications for our understanding of artificial intelligence and its potential impact on society:

1. Redefinition of AGI: The results question conventional assumptions regarding AGI and reveal that AI can perform more than averagely comprehensible cognitive tasks. This calls for reconsidering the concept and the metrics for artificial general intelligence.

2. Educational and Workforce Implications: AI has performed better in knowledge-frontier and understanding-based real-life tasks, leading to fundamental questions for education and the future workforce. With the advancement in AI systems, it is necessary to conceal human tasks and knowledge that are cooperative rather than in conflict with AI systems.

3. Ethical and Societal Considerations: This study's finding of the increasing rate of AI advancement exposes the importance of ethical concerns and policy reviews on emerging technologies. Issues of the rights and responsibilities of AI, as well as a possible shift of people's roles in different fields, must be discussed beforehand.

4. Research and Development Focus: The results imply that the subsequent AI research needs to address not only the efficiency gain on the existing standard tests but also the emergence of new tests and indicators that, in one way or another, reflect the specific aspect of intelligence not included in currently adopted metrics, for instance, emotional intelligence, creativity or abilities that would allow an AI to perform in the conditions that it has not been initially trained for.

## 6.3 Limitations of the Study

While this research provides valuable insights into the current state of AI capabilities, several limitations should be acknowledged:

1. Task Specificity: In this study, the understanding was made of cognitive exercises associated with the knowledge and comprehension of medical functioning. Although these are essential attributes of intelligence, they do not cover all the possible mental abilities of a human.

2. Benchmark Relevance: Therefore, the reliance on educational achievement and literacy as performance indicators can be helpful. However, they are only a part of the higher human characteristics inextricably linked with processes involved in intelligence and problem-solving in real life.

3. Rapidly Evolving Field: Due to the dynamic evolution of the AI field, the performance data of these models are outdated when the study is conducted, which might compromise the long-term comparability of the results made in this research.

4. Lack of Direct Testing: Finally, the study conducted only secondary data analysis instead of directly comparing AI models with actual human participants, and thus may have certain discrepancies in results.

**6.4  Recommendations for Future Research**

Based on the findings and limitations of this study, several avenues for future research are proposed:

1. Comprehensive Intelligence Assessment: Introduce the real and more diverse abilities indicators that could define several Cognitive Skills such as Emotional Intelligence, Creativity, Practical Judgment and Reasoning and Effectiveness in a range of New and Unfamiliar Conditions.

2. Longitudinal Studies: It should be possible to record AI progress and learning over a long time so that the speed at which the technology is developing can be seen and whether there are certain barriers to improving the systems' capabilities.

3. Real-World Application Testing: Conduct practical research in specifying the areas in which it is beneficial to use AI and when human intelligence might perform better in comparison to AI, thus going beyond the approach of comparing AI and humans while solving the existing, well-stipulated tasks that are created specifically for such comparison.

4. Interdisciplinary Approach: Work closely with cognitive scientists, neuroscientists, and philosophers to refine the definitions and metrics of intelligence for use with or for human and artificial entities.

5. Ethical and Societal Impact Studies: Examine possible social consequences of competent AI systems such as employment, learning and social organisation to determine guidelines for policy and usage.

**6.5 Conclusion**

The results presented in the study offer strong evidence that the current large LLMs are already operating at or above the level of the average American in several vital cognitive domains, indicating the significant further steps toward AGI. Nevertheless, such discoveries paint a very optimistic picture of AI and show fundamental improvements in the perceived intelligence level of the algorithms proposed; however, this is also a cause for concern because the observed data emphasize the necessity to reconsider the notion of intelligence and the ways human and artificial intelligence can exist in parallel and manifest themselves. Since there is a position on the brink of a new age in AI, they must persistently analyze and compare these systems while broadening the notion of intelligence to include all the processes indicative of human-level AGI. There are also significant and broad consequences, which means that the constant work of researchers, policymakers, and society, in general, is needed to ensure the successful containment of threats and the use of emerging opportunities provided by the advancements in AI systems.

# References

Anthropic. (2024). The Claude 3 model family: Opus, Sonnet, Haiku. [Technical report].

https://www.anthropic.com/claude3-family-report

Bommasani, R., Hudson, D. A., Adeli, E., Altman, R., Arora, S., von Arx, S., Bernstein, M. S., Bohg, J., Bosselut, A., Brunskill, E., Brynjolfsson, E., Buch, S., Card, D., Castellon, R., Chatterji, N., Chen, A., Creel, K., Davis, J. Q., Demszky, D., ... Liang, P. (2021). On the opportunities and risks of foundation models. arXiv.

https://doi.org/10.48550/arXiv.2108.07258

Bostrom, N. (2014). Superintelligence: Paths, dangers, strategies. Oxford University Press.

Brown, T. B., Mann, B., Ryder, N., Subbiah, M., Kaplan, J., Dhariwal, P., ... & Amodei, D. (2020). Language models are few-shot learners. Advances in Neural Information

Brynjolfsson, E., & McAfee, A. (2014). The second Machine Age: Work, progress, and prosperity in a time of brilliant technologies. W. W. Norton & Company.

Bubeck, S., Chandrasekaran, V., Eldan, R., Gehrke, J., Horvitz, E., Kamar, E., Lee, P., Lee, Y. T., Li, Y., Lundberg, S., Nori, H., Palangi, H., Reif, M., Seltzer, M., & Sinha, K. (2023). Sparks of artificial general intelligence: Early experiments with GPT-4. arXiv.

https://arxiv.org/abs/2303.12712

Goertzel, B. (2014). Artificial general intelligence: Concept, state of the art, and prospects. Journal of Artificial General Intelligence, 5(1), 1–48. https://doi.org/10.2478/jagi-2014-0001

Google Research. (2023). Gemini 1.0 model performance. [Technical report].

https://research.google/pubs/gemini-model-performance/

Hern, A. (2020). AI language model GPT-3 can now generate amazing human-like text – but is it too good? The Guardian. Retrieved from https://www.theguardian.com/technology/2020/jul/22/ai-language-model-gpt-3-openai-amazing-human-like-text

Marcus, G., & Davis, E. (2019). Rebooting AI: Building artificial intelligence the can trust. Pantheon Books.

Muro, M., Maxim, R., & Whiton, J. (2019). Automation and artificial intelligence: How machines are affecting people and places. Brookings Institution. Retrieved from https://www.brookings.edu/research/automation-and-artificial-intelligence-how-machines-affect-people-and-places/

Muro, M., Maxim, R., & Whiton, J. (2019). Automation and artificial intelligence: How machines are affecting people and places. Brookings Institution Press.

National Center for Education Statistics (NCES). (2019). Adult literacy in the United States. Retrieved from https://nces.ed.gov/pubsearch/pubsinfo.asp?pubid=2019179

OpenAI. (2023). GPT-4 technical report. Retrieved from https://www.openai.com/research/gpt-4

Processing Systems, 33, 1877-1901. https://doi.org/10.48550/arXiv.2005.14165

Russell, S., & Norvig, P. (2021). Artificial intelligence: A modern approach (4th ed.). Pearson.

Turing, A. M. (1950). Computing machinery and intelligence. Mind, 59(236), 433–460.

U.S. Census Bureau. (2022). Educational attainment in the United States: 2022. Retrieved from https://www.census.gov/newsroom/press-releases/2022/educational-attainment.html

Vaswani, A., Shazeer, N., Parmar, N., Uszkoreit, J., Jones, L., Gomez, A. N., ... & Polosukhin, I. (2017). Attention is all you need. Advances in Neural Information Processing Systems, 30. https://doi.org/10.48550/arXiv.1706.03762